# NRC-GAMMA: Introducing a Novel Large Gas Meter Image Dataset

Ashkan Ebadi[1,*], Patrick Paul[2], Sofia Auer[2], and Stéphane Tremblay[2]

[1] National Research Council Canada, Montreal, QC H3T 2B2, Canada
[2] National Research Council Canada, Ottawa, ON K1K 2E1, Canada
[*] ashkan.ebadi@nrc-cnrc.gc.ca

**Abstract**

Automatic meter reading technology is not yet widespread. Gas, electricity, or water accumulation meters reading is mostly done manually on-site either by an operator or by the homeowner. In some countries, the operator takes a picture as reading proof to confirm the reading by checking offline with another operator and/or using it as evidence in case of conflicts or complaints. The whole process is time-consuming, expensive, and prone to errors. Automation can optimize and facilitate such labor-intensive and human error-prone processes. With the recent advances in the fields of artificial intelligence and computer vision, automatic meter reading systems are becoming more viable than ever. Motivated by the recent advances in the field of artificial intelligence and inspired by open-source open-access initiatives in the research community, we introduce a novel large benchmark dataset of real-life gas meter images, named the NRC-GAMMA dataset. The data were collected from an Itron 400A diaphragm gas meter on January 20, 2020, between 00:05 am and 11:59 pm. We employed a systematic approach to label the images, validate the labellings, and assure the quality of the annotations. The dataset contains 28,883 images of the entire gas meter along with 57,766 cropped images of the left and the right dial displays. We hope the NRC-GAMMA dataset helps the research community to design and implement accurate, innovative, intelligent, and reproducible automatic gas meter reading solutions.

## 1. Introduction

Natural gas is an economical energy source that is also relatively clean-burning and environment friendly (U.S. Energy Information Administration 2020). For decades, the gas has been supplied to residential and industrial places, being used in many different ways including as a cooking and heating fuel in households, powering industrial furnaces, or even fueling vehicles (King 2013). In Canada, although the gas consumption differs across the country, the main residential use of natural gas is for space and water heating purposes (Government of Canada 2020a). In 2018, Canada consumed an average of 11.2 billion cubic feet per day of natural gas, having Alberta and Ontario provinces as the largest gas consumers (Government of Canada 2020b).

Gas meters are installed to monitor the consumed gas quantity to charge the consumers accordingly. In most countries, the gas meter reading is a manual process being performed by an operator on-site (Deconinck 2010), visually checking the meter and entering the consumption amount manually (Son et al. 2019), and taking a proof image of the meter for the record (Vanetti et al. 2013). The proof image may be later used to verify the reading offline with another operator (Vanetti et al. 2013). Such a method is highly inefficient, time-consuming, expensive, and prone to errors (Laroca et al. 2019; Son et al. 2019). Intelligent automated real-time gas meter reading systems are highly needed to overcome the limitations of the manual process. Such real-time reading systems can be beneficial for both the consumers and the provider companies. From a consumer perspective, it would help them, for instance, to reduce or better manage the consumption and the cost. For a gas provider company, the system would assist them to better understand usage profiles, leaks, and to improve usage predictions.

Although smart meter reading devices are one of the alternative solutions for automatically recording gas (or other utilities) consumption, in many countries, they are not yet widely available and the reading is performed manually (Vanetti et al. 2013). Smart meters can not only record the energy consumption but they can send the recorded data to the supplier for monitoring and billing purposes (Spichkova et al. 2019). Despite clear advantages, this solution is very expensive. For example, the cost of transition from non-smart meters to smart ones in Australia was estimated to be about $ 1.6 billion with a cost burden carried to the customers to pay for the upgrade (Spichkova et al. 2019). Therefore, many countries have delayed the smart meter transition program.

With recent advances in artificial intelligence and computer vision, alternative digital solutions are now more feasible than ever. For example, an automatic meter reading (AMR) system can be designed by installing an image acquisition device, e.g., a camera, in the meter box, and training a machine/deep learning model on the captured real-time images of the meter readings (Edward 2013). Image-based AMRs are cost-efficient, as it is not required to change the meter, and they can be installed quickly (Azeem et al. 2020). They are also scalable and generalizable to different applications and styles of the meters (Edward 2013).

Several studies in the literature proposed/developed either image-based AMR systems or learning methods for a specific task in the system such as text detection in the whole image of the meter or digit segmentation. For example, Yang et al. (2019) proposed a fully convolutional sequence recognition network for the water meter reading. Despite the high performance, the trained model requires the meter-reading region as the input. In another study, Son et al. (2019) proposed an AMR system that contains various components for region detection, digit segmentation, and digit recognition using convolutional neural networks. Lastly, in a very recent study, Azeem et al. (2020) proposed a Mask Region-Convolutional Neural Network (Mask-RCNN) model for automatic meter reading that performs counter detection, digit segmentation, and recognition tasks.

However, most of the AMR-related datasets in the literature are not publicly available as images of the meters often belong to a service company (Laroca et al. 2019). The few existing public meter datasets mostly contain images of digit-based meters, also called counter meters (e.g., Laroca et al. 2019; Yang et al. 2019). This limitation would jeopardize the reproducibility of the performed experiments as well as the possibility of a thorough assessment of proposed systems (Salomon et al. 2020). Additionally, collecting properly and annotating a large amount of data required for training (deep) learning models precisely is an expensive task restricting research individuals/teams with limited financial resources.

To overcome these limitations and inspired by the open-source activities of the scientific communities, we introduce a novel large benchmark image dataset of a residential diaphragm gas meter, called NRC-GAMMA. The NRC-GAMMA dataset is publicly available and contains 28,883 images of the entire gas meter along with 57,766 cropped images of the left and the right dial displays, carefully annotated and systematically validated. To the best of our knowledge, this is the largest open-access dataset of a residential diaphragm gas meter that employed well-defined annotation and validation protocols. The dataset would help researchers and innovators in building deep learning models and/or AI-powered meter reading components/systems to tackle the problem of reading residential meters automatically and accurately with the ultimate goal of proposing intelligent and efficient AMR methods/systems. The rest of the paper proceeds with the "Data" section which describes the gas meter characteristics, the image capturing system that was designed and used, and the NRC-GAMMA dataset properties in detail. The annotation process is introduced in the next section, highlighting the platform used for annotations as well as the mechanisms that were put in place to control and assure the quality of the annotations/labeling. The "Usage Notes" provides users with more information on how to access and use the dataset. The paper concludes in the "Conclusion" section while discussing some challenges and limitations in the final section.

## 2. Data

### 2.1. The Gas Meter Characteristics

The data was collected from an Itron 400A gas meter installed and used in one of the test house facilities of the National Research Council of Canada (NRC), built in 2009 and located in Ottawa, Ontario. The gas meter is a light commercial (also suitable for high-load residential) gas diaphragm meter with a capacity of 400 ft$^3$/hr (Figure 1). The meter has two proving dial displays rotating anticlockwise at different speeds, the left one at 50 dm$^3$ and the right one at 10 dm$^3$ per revolution, and a 5-digit cyclometer display. This type of gas meter is also called an accumulation meter where the counter meter, aka the cyclometer display, is normally used to measure the gas consumption between two readings. That is in practice, with a cyclometer display, dial displays are not used because the time interval between two readings could be too long. We intentionally chose this type of gas meter for at least three reasons: 1) By having continuous monitoring of the dial displays, better gas management for both the consumer and gas providers would be possible, 2) A well-annotated large-scale dial display meter dataset is not publicly available to the best of our knowledge, and 3) There are other types of utility in addition to the gas meters that only have dial displays. Having a labeled set of dial displays would expand the application of the NRC-GAMMA dataset to other utility meters as well.

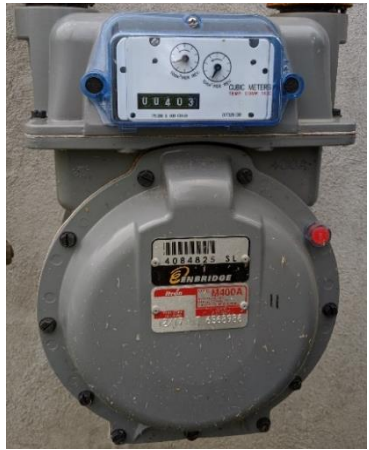

**Figure 1.** The Itron 400A gas diaphragm meter.

### 2.2. The Image Capturing System

To ensure that the images are captured continuously and consistently, we built a smart image capturing system powered by advanced edge processing techniques able to capture images in different conditions during day and night (Figure 2-a and -b). The system contained a Raspberry Pi 3b+, i.e., a small single-board computer, powered by a Power over Ethernet (PoE) HAT that allows powering the Raspberry Pi board using PoE–enabled networks, a 5-megapixel infrared camera module (NoIR) that enabled us to take quality pictures at night as well using infrared lighting. The components were all boxed in a weatherproof openH Rubicon IP67 case. The images were taken automatically on average every ~3 seconds on the 20$^{th}$ of January 2020 between 0:05 and 23:59 Eastern Time, with the average temperature of 16.5 °C and wind blowing with the average speed of 11 km/hour. Figure 2-c and -d depict sample images of the meter region captured by the system during daylight and at night, respectively.

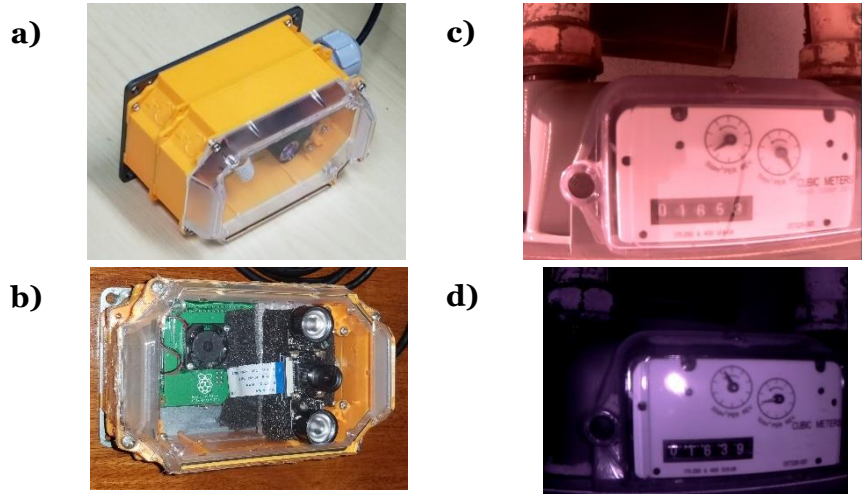

**Figure 2**. **a**) The image capturing system, side view, **b**) front view, **c**) sample day time captured image of the meter, and **d**) sample image of the gas meter captured at night.

### 2.3. The NRC-GAMMA Data Set Characteristics

The NRC-GAMMA data set contains 28,883 raw images of the entire gas meter along with cropping metadata and script to create 57,766 cropped images of proving dials from the original images, i.e., 28,883 images per dial. That is the data set includes two types of images: 1) the raw images of the gas meter, and 2) cropped images of the left and right dials, separately. The cropped images (160 * 160 pixels) only show the region of interest, i.e., only the dial region, and were included to facilitate researchers' analytic efforts in case they would like to only analyze the dial region rather than the whole image of the gas meter. Figure 3 shows a sample gas meter image along with cropped regions of the left and right dials.

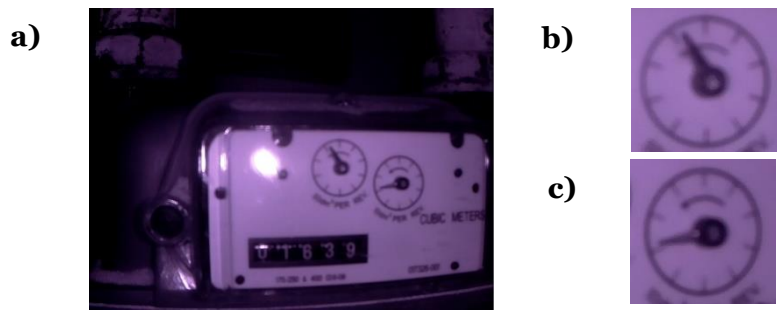

**Figure 3. a**) A sample raw image of the gas meter panel captured at night, **b**) cropped image of the respective left dial, and **c**) cropped image of the respective right dial.

Since the images were taken during the whole day, the camera was affected by various environmental factors, e.g., varying lighting conditions and wind speed that resulted in various sets of artifacts in the captured images. We intentionally created this situation since these diverse variations in the dataset provide researchers with a real-life example allowing them to stand for variations and create flexible and reliable AMR systems that fit real-life conditions. Figure 4 shows a sample image with no artifact plus some sample captured images with various observed artifacts such as shadow and light reflection.

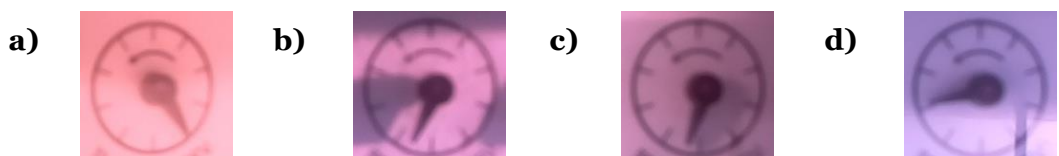

**Figure 4. a**) A sample cropped image with no artifact, **b**) with heavy shadow, **c**) blurry due to windy condition, and **d)** glare due to direct sunbeam reflection.

## 3. The Annotation Process

### 3.1. Amazon Mechanical Turk as the Annotation Service Provider

To annotate and label images in the NRC-GAMMA data set, we used Amazon Mechanical Turk (MTurk) which is a crowdsourcing service to perform discrete on-demand tasks. We created a user-friendly interface (Figure 5) facilitating the annotation process for annotators while decreasing the margin of error by providing an easy-to-use graphical user interface (GUI). Given a cropped image of a dial display, the annotator was tasked to click on the respective region in the interface that in their opinion was the most representative annotation for the image. To avoid ambiguity and to follow a consistent process, we instructed annotators to take the next region (anticlockwise) as the label in cases where they were uncertain between two regions. We created random batches of 1,000 images of the dials, i.e., 58 batches in total. Each image was annotated by at least 3 annotators. A large proportion of the images were annotated by 6 annotators. In the next sections, we explain in detail the annotation process and the systematic approach that we followed to set the final labels for the images and to ensure the quality of the labels.

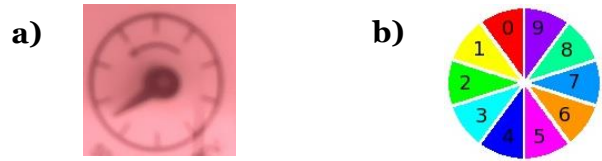

**Figure 5.** A sample annotator's task that contained two parts, **a)** a sample dial image, and **b)** the interface in which the annotator should select the representative region based on the given dial image. In this example, the correct answer is 3.

### 3.2. High-Level Annotation Process

Our defined annotation process accounts for subjectivity and human error and reduces their impact to a minimum. Figure 6 shows a high-level conceptual flow of the decision logic. The high-level annotation logic for a given dial image is based on two iterations and is described as follows:

1. In the first iteration, the given dial image is sent to three distinct annotators.
    a. When all three annotations have unanimous consensus, i.e., three annotations are identical, it is set as the label of the image and the process is terminated for the given image.
    b. Otherwise, the annotations are kept as *Set-1* and we go to *Iteration 2*.
2. In the second iteration, the image is sent to three new distinct annotators.
    a. If all three annotations received from the new annotators are unanimous, the annotation is set as the label of the image, and the process is terminated.
    b. Otherwise,
        i. The new annotations are stored as *Set-2*.
        ii. *Set-1* and *Set-2* are merged and named as the *annotations set*.
        iii. If there is a unique majority mode (without ties) in the *annotations set*, then this mode is set as the label, and the process is terminated.
        iv. If the given image has a tie in the mode, whereby the two modes are adjacent, then the image is labelled with the 2 adjacent modes, and the process is terminated.

v. All other images not labelled at this point, are visually inspected and labelled accordingly.

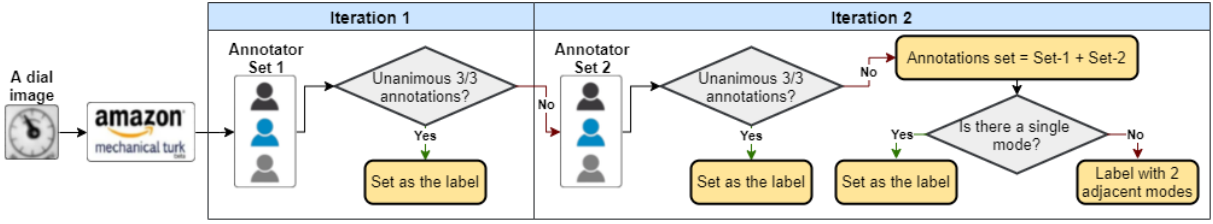

**Figure 6.** The high-level conceptual flow of the decision logic for setting labels of the gas meter dial images.

### 3.3. Annotation Methodology and Quality Control

We systematically monitored and verified the quality of annotators and annotations during the labelling procedure. The quality control and data integrity checks were performed by coding scripts in the R programming language. The R scripts are available upon request. In this section, we explain the annotation methodology in detail.

### 3.3.1. Quality of Annotators

We understand the annotation difficulty level varies for various reasons such as with the position of the hand in the dial display or the time of day. Evaluating annotators is not simple. We assessed the quality of annotators, after receiving the first iteration of annotations. First, we define an "*unacceptable*" annotation if one of the following two conditions occurs: 1) it is an annotation for an image where there is no consensus or majority in the annotation set, e.g., {4, 8, 3}, or 2) it is an annotation related to an image where there is a majority mode, and it is the annotation that did not make up that majority, and also is not adjacent to the majority mode value for that image. For instance, if the 3 annotations of an image are {1, 1, 3}, the 3 is considered unacceptable as it is not the majority mode and it is not adjacent to the majority mode, i.e., +/- 1 of the majority mode of 1. Had the last annotation been a 0 or a 2 (instead of the 3) then that annotation would have been considered acceptable. Second, the total number of annotations completed by each annotator was calculated, and the quality of an annotator was measured using the following metric:

$$Q_a = \frac{A_u}{A_t} \quad (1)$$

where $Q_a$ is the quality of a given annotator, $A_u$ is the total number of unacceptable annotations, and $A_t$ is the total number of annotations completed by an annotator. We grouped the annotators based on the quality of their annotations (i.e., their $Q_a$) into three categories: 1) high quality, if $Q_a < 0.2$, 2) fair, if $0.2 \leq Q_a \leq 0.5$, and 3) low quality, if $Q_a > 0.5$. In other words, an annotator with low quality of work will assign more than half of its images with an unacceptable label. In those cases, we preferred to not set the label solely based on its annotations and considered another set/round of annotation to make the decision.

We constantly monitored and evaluated the quality of annotators during the annotation process. Of the 57,766 images sent for the first round of annotations, 1,102 annotators produced 173,298 annotations, of which 10,026 annotations (6%) were considered unacceptable. Among those annotators, 10% were classified as low quality, and 90% were annotators with fair or high quality of work (5% and 85% respectively).

### 3.3.2. Detailed Annotation Methodology

Figure 7 shows the detailed annotation methodology with the corresponding number of images at each step. As seen, the process started with 57,766 cropped images of the left and right dial displays (28,883 images per dial). In the first round of annotation, images were sent to three annotators in the Mechanical Turk platform and three annotations were received for each image, i.e., 173,298 annotations in total. From 57,766 images, 37,981 images (65.7%) were annotated with a unanimous label that is all three annotators agreed on the label. For the rest, i.e., 19,785 images, the annotations were not unanimous that is there was at least one disagreement.

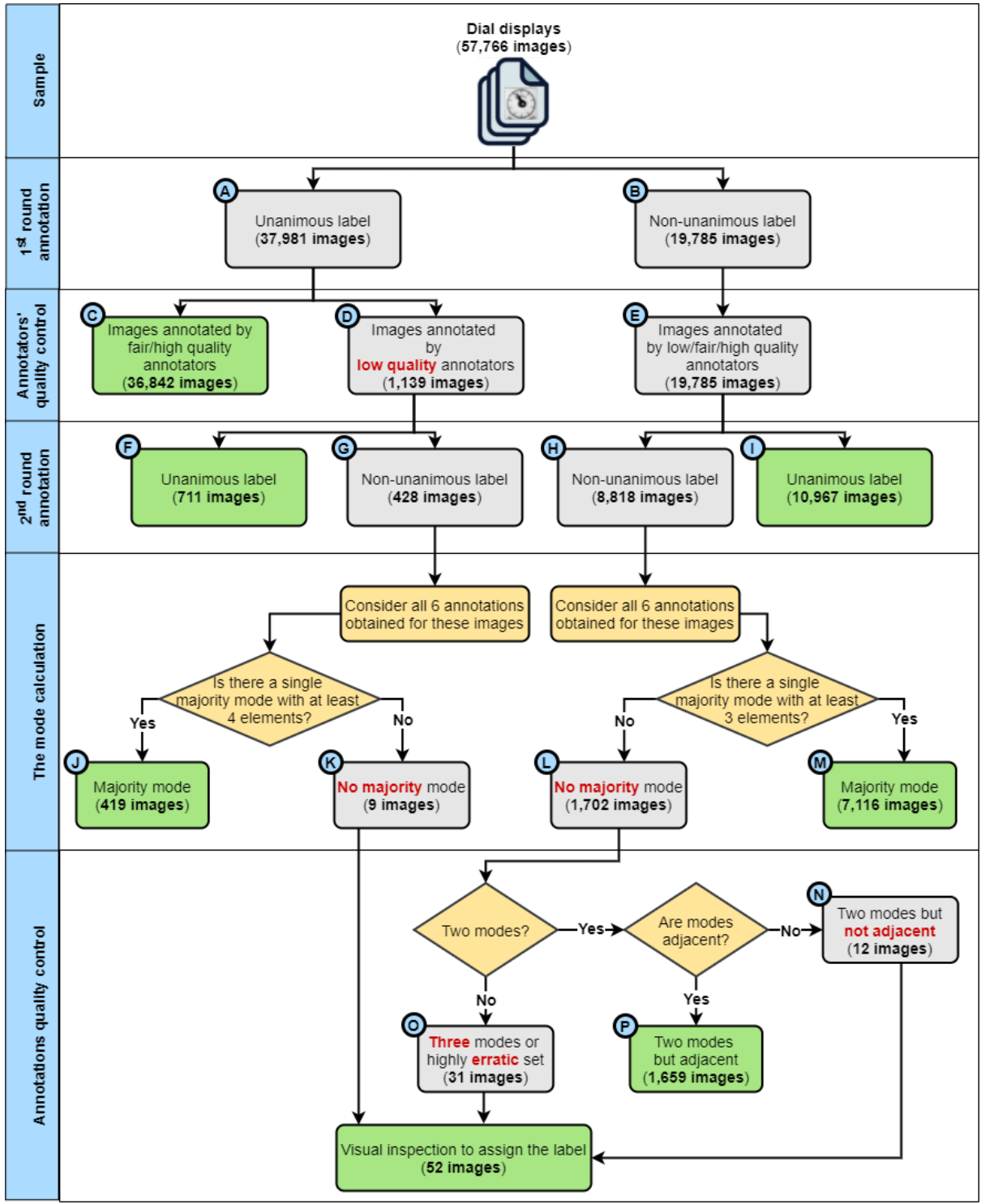

**Figure 7.** Detailed annotation methodology with the corresponding number of images at each step.

In the next step, we checked the quality of annotators using the quality metric defined in the previous section ($Q_a$). For the images with a unanimous label in the first round of annotation ($n = 37,981$, left branch in Figure 7, box A), 36,842 images (97%, box C in the figure) were annotated by annotators with fair/high quality ($Q_a \leq 0.5$). The unanimous label was set and the annotation process terminated for this set of images. For the remaining 1,139 images (3%, box D in the figure) with unanimous label quality of at least one annotator was not satisfactory, hence, the set remained in the process. Images with a non-unanimous label in the first round of annotation ($n = 19,785$, right branch in Figure 7, box B), were also considered for further verification and remained in the process, regardless of the quality of annotators (box E in Figure 7), however, we restricted annotators with low/fair quality from annotating again.

We did a second round of annotation for the remaining images, i.e., all 20,924 images that did not have unanimous annotations (19,785, box E in Figure 7) or those from low-quality annotators (1,139, box D). For images annotated by at least one annotator with low quality in the first round of annotation (box D), if we obtained a unanimous label in the second round, it was set as the label and the annotation process terminated for this set of images ($n = 711$, box F in Figure 7). We kept the remaining images ($n = 428$, box G) in the process for further verification. We followed the same process for images with a non-unanimous label in the first round of annotation (box E). Out of these 19,785 images, a unanimous label obtained in the second round of annotation for 10,967 images (55%), it was set as the label, and the annotation process terminated for this set of images (box I in the figure). The remaining images ($n = 8,818$, box H) were kept in the process. Up to this step, a label was set for 48,520 images (84%).

We followed a careful quantitative/manual process for the remaining 9,246 images (16%) to set the label. For these images (boxes G and H in Figure 7), we first combined the annotations received in both rounds, i.e., 3+3 annotations per image. For images with non-unanimous annotation in the second round and unanimous annotation but at least with one low-quality annotator in the first round of annotation (box G in Figure 7), we checked if there exists a single, unique, no tie, majority mode with at least 4 elements in the set of 6 annotations. If such mode was found, it was assigned as the label for those images, and the process terminated ($n = 419$, box J in Figure 7). For images with a non-unanimous annotation in both rounds of annotation (box H in Figure 7), we checked if there exists a single, unique, no tie, majority mode with at least 3 elements in the set of 6 annotations. If such mode was found, it was assigned as the label for those images, and the process terminated ($n = 7,116$, box M in Figure 7). We applied a stricter rule for images in box G compared with those in box H (i.e., mode with at least 4 elements vs. mode with at least 3 elements) as we identified some low-quality annotators in the first round of annotation for images in box G despite agreeing on a unanimous label. After this step, a label was set for 56,055 images (97%).

We did further verification for the remaining 1,711 images (3%). Those 9 images in the left branch of the decision tree with no majority mode (box K in Figure 7) were sent for visual inspection. For 1,702 images in box L in the figure, we first checked each image if there exist two modes in the set of 6 annotations. If yes, we checked if those two modes refer to adjacent zones in the dial display, and if they were adjacent, we duplicated those images and assigned one of the modes to each of the duplicated images, and terminated the process for them ($n = 1,659$, box P in Figure 7). If the two modes were not adjacent, we sent those images for visual inspection ($n = 12$, box N in Figure 7). The reason for the adjacency condition was to ensure that the main cause of disagreement between annotators is the difficulty level of the image and not a careless/random annotation. Also, if there were more than two modes, or there is one mode but the annotation set is highly erratic that is there are so many other values (e.g., {2, 2, 1, 3, 5, 9}), or no mode at all in the set of 6 annotations ($n = 31$, box O in Figure 7), the image was sent for visual inspection. Overall, 52 images were sent for visual inspection (from boxes K, N, O in Figure 7). In the visual inspection step, images were annotated by three subject experts from our team, and the majority voting rule was applied to assign the final label.

### 3.4. Data Records

The NRC-GAMMA benchmark dataset is available to the general public at https://github.com/nrc-cnrc/NRC-GAMMA. The raw and cropped images are located at the NRC Digital Repository in compressed TAR.GZ archive file format. The current version of the data set contains 28,883 raw images of the entire gas meter along with 57,766 cropped images of dial displays, 28,883 for left and right dials, respectively. We defined 2 fixed regions of interest (ROI) on each original image of the gas meter to create the cropped images for the left and right dial displays as follows: 1) for the left dial: a rectangle from ($y_1$=325:$x_1$=485) as the top left corner to ($y_2$=552:$x_2$=712) as the bottom right corner, and 2) for the right dial: a rectangle from ($y_1$=365:$x_1$=525) to ($y_2$=703:$x_2$=863). The cropped images of the left and right displays are of 160 * 160 pixels dimension. Users may use different cropping approaches to extract their ROIs of interest from the original images of the gas meter based on their use case and/or research objectives. This makes the NRC-GAMMA data set highly flexible for various research objectives. Each cropped image represents a single snapshot of the left/right dial display of the gas meter with an associated timestamp, dial indicator, and the respective label in the filename. For example, "L" in "L_1579496405557_0.jpg" filename indicates that the image is of the left dial display. The middle part, i.e., "1579496405557", shows the captured integer timestamp that is the number of seconds since the 1$^{st}$ of January, 1970, 12:00:00am (UTC). Using the timestamp in the filenames as the key, users can link the cropped images back to the original images of the entire gas meter. The last part in the filename, i.e., "0" in the example, indicates the label of the given image.

Table 1 lists the location and description of files included in the NRC-GAMMA archive file. The original images of the entire gas meter are located in the "/data/original" folder. The cropped images of the left and right dial displays are located in the "data/cropped" folder. Metadata of the cropped images are provided in the 'Metadata.csv' file under the root directory in the archive file. The ReadMe.md file provides instructions on using the dataset, information and policies on licensing and distribution of the dataset, as well as a brief description of all the files included in the NRC-GAMMA archive file.

**Table 1.** Description and location of the files included in the NRC-GAMMA archive file.

| File | Location | Description |
|---|---|---|
| Original images | "data/original" | This folder contains the original images of the entire gas meter. |
| Cropped images of the left and right dial displays | "data/cropped" | This folder contains the cropped images of the left and right dial displays. |
| Metadata.csv | Root directory | Metadata on the gas meter images. |
| ReadMe.md | Root directory | Includes instructions to use the dataset, information on licensing and distribution, and a summary of the included files and folders. |

### 4. Usage Notes

The NRC-GAMMA GitHub repository includes metadata, documentation, instructions to use, and scripts to download the TAR.GZ archive files. The NRC-GAMMA is an evolving data set.

We are constantly searching for improvements and even more data, therefore, the NRC-GAMMA data set and metadata will be evolving/growing over time. We recommend that users check the NRC-GAMMA repository at https://github.com/nrc-cnrc/NRC-GAMMA, for the latest versions of data, metadata, and scripts. In the GitHub repository, users are provided with a Python notebook that contains all the steps required to download the data set archive file, extract the original gas meter images as well as the cropped ones, and store them on their local device. The provided scripts are well-documented allowing users to modify the script, e.g., use different cropping parameters, based on their research objectives and requirements if needed.

## 5. Conclusion

Smart gas meters enable automatic reading of gas consumption, however, they are not yet being widely deployed even in the developed countries, having many conventional gas meters in operation. The recent advances in the fields of artificial intelligence and computer vision offer alternative solutions through intelligent image-based automatic gas meter reading systems. Such systems require huge datasets of gas meter images to train their model on. However, such image datasets are rarely publicly available as the meter images often belong to a service company. Moreover, the few existing public gas meter datasets mostly contain images of the counter meters and not the dials. This limitation would jeopardize the reproducibility and scalability of the proposed automatic gas meter reading systems. Additionally, collecting and annotating a large amount of data is an expensive task restricting research individuals/teams with limited financial resources.

To overcome such limitations and in an effort toward supporting the scientific community through open-source open-access initiatives, we introduced a large labeled public benchmark image dataset of residential diaphragm gas meter readings, named the NRC-GAMMA. The dataset contains 28,833 images of the entire gas meter along with 57,766 cropped images of the left and right dial displays, 28,883 images per dial, that were systematically annotated, carefully validated, and intensively monitored/controlled to ensure the quality. We hope the NRC-GAMMA dataset helps the research community in building accurate intelligent solutions to tackle the problem of reading residential meters automatically. This would optimize the currently expensive, labor-intensive, and error-prone manual reading of the gas meters.

## 6. Challenges and Limitations

Image annotation is a challenging, time-consuming, and labor-intensive task and requires a well-defined process to reduce the error margin. Quality of annotation directly affects the quality of the AI models trained on those images (Cheng et al. 2018) as any AI model is only as good as the data it was trained on. Hence, in NRC-GAMMA, we checked for the quality of annotations at multiple levels/steps to ensure that the data that will be fed into the algorithms are of the best quality. The other challenge we overcame was designing/building an annotation tool/interface to facilitate the annotation process for the annotators. Such a tailored easy-to-use interface helped us to reduce human error further.

Despite employing a well-documented and careful process, like any process, our annotation pipeline has some limitations. We used two layers of annotation, each involving various sets of annotators, with strict control on the performance of annotators and the quality of their annotations in the first layer. We did not apply that strict annotators' quality control in the second layer of the annotation process as in that case the process could involve many more annotation loops, taking a lot of time and requiring additional budget. We rather merged annotations received in both rounds of annotation for those images, applied a threshold-based mode calculation to assign the label, and visually checked the extremely hard examples to ensure the quality. As part of the process, we ended up with 1,659 images with 2 labels and decided to duplicate them, having one image per label. We randomly checked a few of these images and found them extremely hard to annotate as the hand in the dial display was often right on the tick. We intentionally included these images to enable research scenarios/questions that would like to focus on these extreme cases. Users could also exclude them from their analyses if they find them not useful for their research objectives.

## Acknowledgment

We would like to offer our special thanks to Mr. Tyson Mitchell (National Research Council Canada) and Ms. Lama Elnaggar for their precious help and support during this project.